\relax
\documentclass[letterpaper]{article}
\usepackage{ijcai16}
\usepackage{times}
\usepackage{helvet}
\usepackage{courier}
\usepackage{graphicx}
\usepackage{amsmath}
\usepackage{multirow}
\usepackage{booktabs}
\usepackage{array}
\usepackage{subfigure}
\usepackage{amssymb}
\usepackage{CJK}

\begin{document}
%
\title{Neural Emoji Recommendation in Dialogue Systems}
\author{Ruobing Xie$^{1}$, Zhiyuan Liu$^{1}$, Rui Yan$^{2}$, Maosong Sun$^{1}$\\
$^{1}$ Department of Computer Science and Technology, \\
State Key Lab on Intelligent Technology and Systems, \\
National Lab for Information Science and Technology, Tsinghua University, Beijing, China \\
$^{2}$ Natural Language Processing Department, Baidu Inc., China \\
}

\maketitle
\begin{abstract}
Emoji is an essential component in dialogues which has been broadly utilized on almost all social platforms. It could express more delicate feelings beyond plain texts and thus smooth the communications between users, making dialogue systems more anthropomorphic and vivid. In this paper, we focus on automatically recommending appropriate emojis given the contextual information in multi-turn dialogue systems, where the challenges locate in understanding the whole conversations. More specifically, we propose the hierarchical long short-term memory model (H-LSTM) to construct dialogue representations, followed by a softmax classifier for emoji classification. We evaluate our models on the task of emoji classification in a real-world dataset, with some further explorations on parameter sensitivity and case study. Experimental results demonstrate that our method achieves the best performances on all evaluation metrics. It indicates that our method could well capture the contextual information and emotion flow in dialogues, which is significant for emoji recommendation.
\end{abstract}

\section{Introduction}

Emojis, which are some graphic symbols or small pictures expressing our feelings and emotions, are widely loved and utilized by large amounts of users in almost all social platforms such as Twitter, Facebook and Weibo. The bloom of Emojis has changed conventional communication schemes that only use plain texts, making the conversations between two speakers much more vivid and interesting. Moreover, emojis are informative and flexible that could even express some profound meanings beyond words and sentences.

\begin{figure}[!htbp]
\centering
\includegraphics[width=0.90\columnwidth]{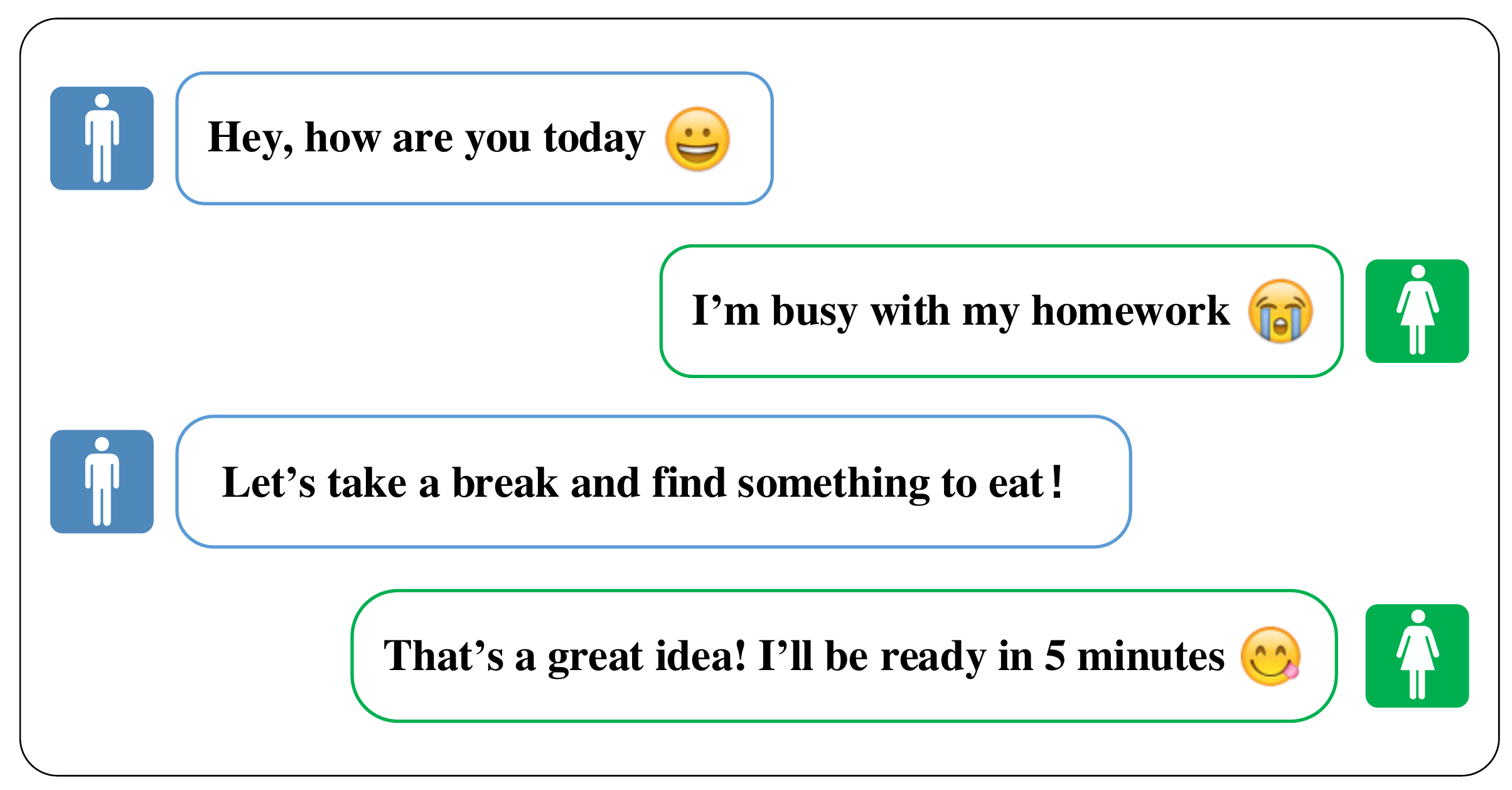}
\caption{An example of emojis in a dialogue.}\label{fig. 1}
\end{figure}

With the thriving of emojis appearing everywhere in our daily lives, the user preferences and behaviours behind emojis have attracted great attention recently. Specifically in dialogue systems, we are supposed to generate the appropriate replies according to the posts given by users, and those replies will be more lively and anthropomorphic if combined with emojis. Fig. \ref{fig. 1} demonstrates an example of emojis utilized in a dialogue. With the favor of emojis, users can smoothly express their feelings of grief and happiness through concise sentences on social platforms.

In this paper, we aim to automatically recommend appropriate emojis attached to the current reply in multi-turn dialogue system according to the contextual information. Emojis could make the generated replies more anthropomorphic and interesting, which will significantly enhance user experiences when using dialogue systems. It is intuitive to recommend emojis according to the reply sentences directly. However, since the meanings of sentences in multi-turn dialogues strongly depend on their contexts, simply considering the reply sentences will not fully understand the whole dialogues. For instance, the last reply in Fig. \ref{fig. 1} is followed by a \emph{delicious} emoji, while the implication of delicious food only appears in other sentences. We believe firmly that the multi-turn contextual information in dialogues should be well considered for emoji recommendation.

We formalize this task as emoji classification. Given a dialogue, we first attempt to understand the whole conversation's meaning, and then predict appropriate emoji(s) for the current reply. This classification task seems to be similar with sentiment analysis, while the differences between this two tasks are still significant: (1) sentiment analysis typically focuses on predicting the sentiment polarities of sentences or documents, while emoji classification attempts to recommend from larger amounts of candidates, which are much more detailed and complicated to analyze. (2) In sentiment analysis, the sentiment polarities are relatively objective and stable. On the contrary, the usages of emojis in real-world dialogue systems are rather subjective and flexible, significantly influenced by user preferences and specific scenarios. Both differences increase the uncertainty in emoji classification, and thus make this task more challenging.

To fully understand the emotions in dialogues, we propose our emoji classification framework taking multi-turn dialogue information into consideration. Impressed by the power of deep learning, we implement several neural network models to learn the dialogue representations and then classify with the learned features. We evaluate our models as well as baselines on a real-world dialogue dataset, and explore some further analysis on parameter sensitivity and representative cases. Experimental results demonstrate that our methods could understand the latent contextual information located in dialogues well, and are capable of generating appropriate emojis attached to the reply sentences. The main contributions of this paper are shown as follows:
\begin{itemize}
  \item To the best of our knowledge, we are the first to introduce rich contextual information to emoji classification task in multi-turn dialogue system.
  \item We propose a novel hierarchical long short-term memory model to construct better dialogue representations, which achieves the best classification performance.
  \item We evaluate our models on a real-world dialogue dataset and do some further analysis on representative cases.
\end{itemize}

\section{Related Work}

Recent years have witnessed the widespread usage of emojis on computer-mediated communication. Emojis or emoticons could better express user emotions beyond plain texts, making communications more lively and smoother. Experimental results indicate that users are more satisfied with communications with emoticons \cite{rivera1996effects}, and emojis indeed help users to exchange emotions and thus enhance the message content \cite{yigit2005emoticon}. To utilize the rich information in emojis, \cite{go2009twitter} proposes a distant supervision approach for Twitter sentiment classification with emojis regarded as noisy labels. \cite{liu2012emoticon,zhao2012moodlens,kiritchenko2014sentiment} also consider emojis as significant features in sentiment analysis. However, the usage of emojis is flexible and rather individual, which may provoke differential effects according to different person perception patterns \cite{ganster2012same}. There are large amounts of works focusing on interpreting the differences in emoji usage across genders \cite{wolf2000emotional} or cultures \cite{park2013emoticon}. The flexibility in emojis makes communications more vivid, while it also makes the emoji classification much more challenging.

Emoji classification could be inspired by the previous work on sentiment analysis, for these two tasks both attempt to understand the emotions of inputs and give appropriate predictions. Socher introduces recursive autoencoder \cite{socher2011semi} and recursive neural tensor network \cite{socher2013recursive} to sentiment analysis, while \cite{tang2015document} utilizes gated recurrent neural network to learn document representations with their hierarchical structures. \cite{yang2016hierarchical,chen2016neural} further adopt attention on more informative words and sentences with the helps of internal and external information, which significantly improves the performances on sentiment analysis. These explorations on sentiment analysis could inspire the task of emoji classification, while there are still large gaps between these two tasks due to the differences in input forms and classification complexity. To the best of our knowledge, our model is the first attempt on emoji classification by taking the contextual information into consideration in multi-turn dialogue systems.

We attempt to utilize neural networks to learn dialogue representations, among which the recurrent neural network (RNN) is naturally fit for encoding sequential inputs. \cite{hochreiter1997long} proposes the long short-term memory network (LSTM) which aims to address the problem of gradient vanishing in RNN, while \cite{gers2000learning} polishes the original LSTM model by introducing the forget gate. Recently, with the thrives in deep learning, the LSTM models have been widely utilized in various fields such as machine translation \cite{sutskever2014sequence}, natural language generation \cite{li2015hierarchical} and machine reading \cite{liu2016sentence}. Inspired by the great successes in using LSTM, we introduce the hierarchical LSTM model to emoji classification for better understanding multi-turn dialogues.

\section{Methodology}

We first introduce the notations utilized in this paper. Let $D=\{d_1, d_2, \cdots, d_m\}$ denotes the overall dialogue training set, with $m$ considered as the number of dialogues. A dialogue $d$ is usually comprised of a sequence of sentences between two speakers, which we represent as $d=\{s_1, s_2, \cdots , s_{n_d}\}$, with $n_d$ considered as the number of sentences of this dialogue. The last sentence $s_{n_d}$ in a dialogue is considered as the \textbf{reply sentence} for which we should generate emojis. For a sentence $s=\{x_1, x_2, \cdots, x_{n_s}\}$, $x_i \in X$ stands for the words and $n_s$ is the length of this sentence. We also set the emoji set $E=\{e_1, e_2, \cdots, e_{n_e}\}$. For each dialogue instance in $D$, there must be at least one emoji appearing in the reply sentence.

\subsection{Overall Architecture}

\begin{figure}[!htbp]
\centering
\includegraphics[width=0.85\columnwidth]{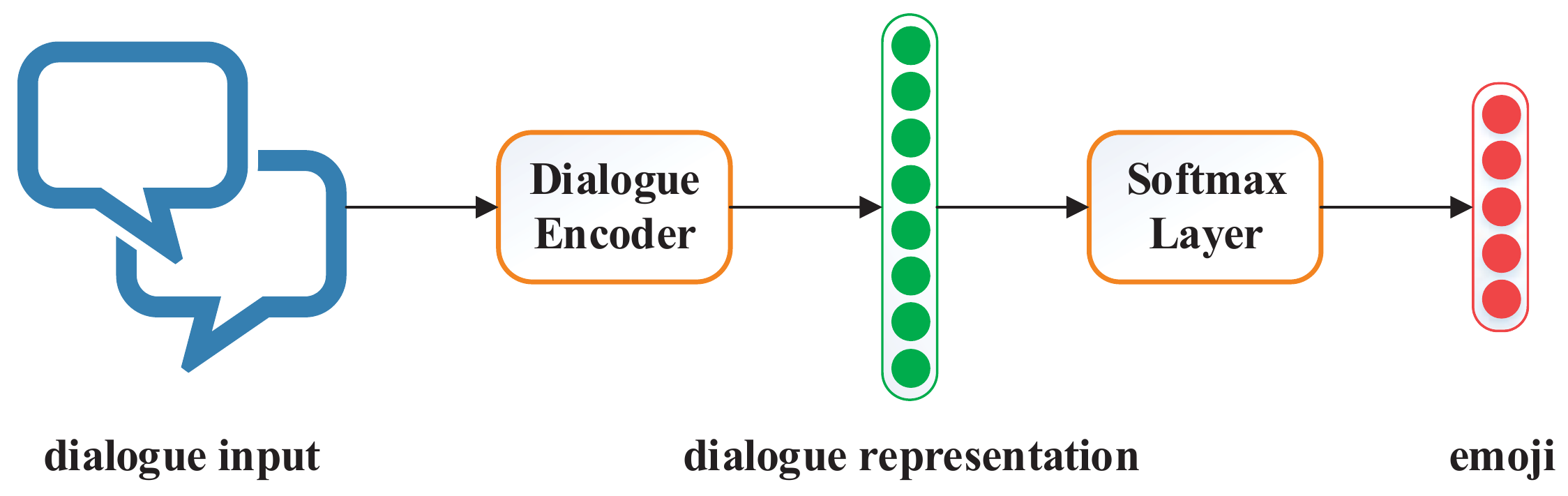}
\caption{Overall architecture of our model.}\label{fig. 2}
\end{figure}

Emoji classification aims to give appropriate emojis attached to the reply sentences according to the contextual information in dialogues. The overall architecture of our model is demonstrated in Fig. \ref{fig. 2}. First, all dialogue instances are considered as inputs after data preprocessing. Second, we design two neural dialogue encoders to construct the dialogue representations, aiming to extract informative and discriminative features located in plain texts. Finally, we utilize a softmax classifier to calculate the probabilities of all emoji candidates and then give the most appropriate predictions.

\subsection{Input Representation and Preprocessing}

\begin{figure*}[!htbp]
\centering
\includegraphics[width=0.86\textwidth]{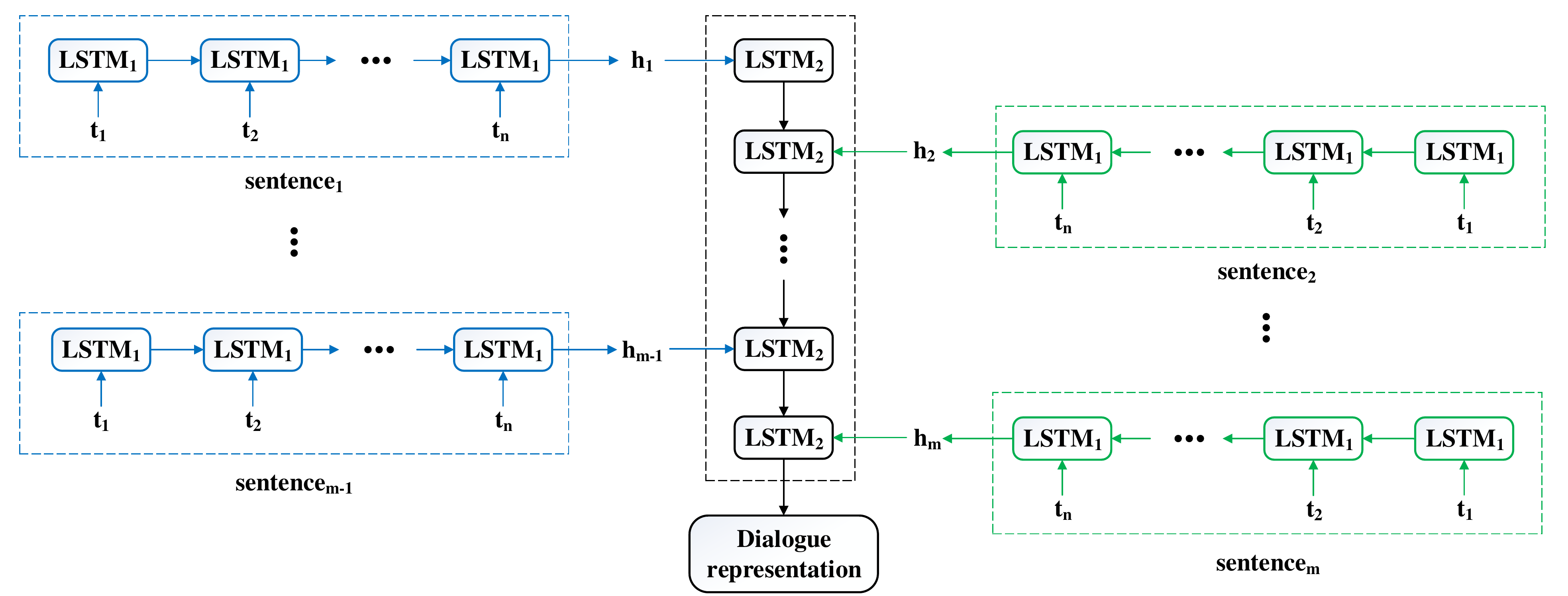}
\caption{Hierarchical long short-term memory in dialogue system.}\label{fig. 3}
\end{figure*}

A dialogue $d=\{s_1, s_2, \cdots , s_{n_d}\}$ is the conversation between two speakers, which could be easily represented as a sentence sequence. Also, a sentence $s=\{x_1, x_2, \cdots, x_{n_s}\}$ could be represented as a word sequence. Note that the ``word" here consists of not only the general word, but also other symbols. Each word embedding is projected into a low-dimensional vector space which takes value in $\mathbb{R}^{m_x}$. These two different granularities of sequences build the hierarchical structure in dialogues.

In preprocessing, we first decide which emojis will be utilized in our classification system, and then extract the dialogues which contain any of those emojis from the original datasets. Next, for each dialogue, we take one sentence containing the selected emojis as the reply sentence if there are multiple candidates. To simulate the real-world scenario in dialogue system, we only consider the conversations before the reply sentences when predicting. The dialogue and sentence lengths are also limited to avoid the possible mistakes in modeling long-term memories. Finally, date cleaning and word segmentation are implemented if necessary.

\subsection{Dialogue Encoder}

The dialogue encoder takes each dialogue after data preprocessing as input, and attempts to learn the compressed low-dimensional representation for each dialogue. The dialogue representations will be then utilized for emoji classification via softmax classifier. In this section, we propose three dialogue encoders to learn dialogue representations.

\subsubsection{Single Long Short-Term Memory}

The single long short-term memory (S-LSTM) is a basic LSTM model which merely considers the reply sentences as inputs. It is natural and straightforward to recommend emojis only focusing on the reply sentences themselves. The LSTM model has three gates, namely the input gate, the forget gate and the output gate. With the favor of multiple gates, LSTM models could well control the information flow in sentences to learn and forget things intelligently. To address the issue of gradient vanishing, LSTM also maintains a memory cell for long-distance dependencies. Suppose the reply sentence is written as $s=\{x_1, x_2, \cdots, x_{n}\}$, the last hidden state is then considered to be the dialogue representation.

\subsubsection{Flattened Long Short-Term Memory}

The S-LSTM model only concentrates on the reply sentences, regardless of the rich information in the contexts, which may harm the performance of recommendation. Conversations in real-world dialogue systems are usually involved with omissions and intimations, whose meanings can only be detected from their contextual information. To take advantages of multi-turn dialogues, we propose the flattened long short-term memory (F-LSTM), which concatenates all sentences in each dialogue sequentially to form a long sequence. Specifically, we define $s_i=\{x_{i1}, x_{i2}, \cdots, x_{in}\}$ as the $i$-th sentence in dialogue, and the input sequence of F-LSTM after flattening will be $\{x_{11}, \cdots, x_{1n}, \cdots, x_{m1}, \cdots, x_{mn}\}$.

\subsubsection{Hierarchical Long Short-Term Memory}

Although the F-LSTM model utilizes the whole dialogue to construct dialogue representations, simply considering the input dialogue as a long word sequence will lose the hierarchical structures when adapting to the input form of LSTM models. The flattening operation breaks the hierarchical structures in dialogues, and it is still hard for the LSTM model to memorize all informative messages if the dependencies are too long.

Inspired by \cite{li2015hierarchical}, we utilize the hierarchical long short-term memory (H-LSTM) model to alleviate the limitations caused by standard LSTM models. Fig. \ref{fig. 3} demonstrates the overall architecture of H-LSTM. Differing from the standard LSTM, the H-LSTM model analyzes input dialogues hierarchically. More specifically, in the word layer, we first utilize a shared LSTM model to learn each sentence representation separately. We have:
\begin{equation}
\begin{split}
\mathbf{h}_t^{(1)}=LSTM_1(\mathbf{x}_t,\mathbf{h}_{t-1}^{(1)}).
\end{split}
\end{equation}
The last hidden state $\mathbf{h}_{n_{i}}^{(1)}$ for the $i$-th sentence is regarded as the corresponding sentence representation. Next, in the sentence layer, we utilize another LSTM model taking these sentence representations sequentially as inputs to generate the overall dialogue representation. The sentence-level LSTM is then formalized as follows:
\begin{equation}
\begin{split}
\mathbf{h}_t^{(2)}=LSTM_2(\mathbf{h}_{n_{t}}^{(1)},\mathbf{h}_{t-1}^{(2)}).
\end{split}
\end{equation}
The last hidden state $\mathbf{h}_{n_d}^{(2)}$ in the sentence level is considered as the dialogue representation $\mathbf{d}$. In H-LSTM, we first attempt to learn each sentence's meaning, and then further understand the whole dialogue through all sentences, which is exactly what human do in real-world conversations.

\subsection{Objective Formalization}

Once we get the dialogue representations from dialogue encoders, the next step is to predict appropriate emojis that should appear in the reply sentences. We utilize a softmax layer for classification as follows:
\begin{equation}
\begin{split}
p(e_i|d)=\frac{\exp(\mathbf{W}_{s_i}\mathbf{d}+\mathbf{b}_{s_i})}{\sum_{j=1}^{n_e}\exp(\mathbf{W}_{s_j}\mathbf{d}+\mathbf{b}_{s_j})},
\end{split}
\end{equation}
in which $p(e_i|d)$ stands for the probability of $i$-th emoji given the dialogue $d$. $\mathbf{W}_s$ is a projection matrix and $\mathbf{b}_s$ is the bias. We use cross-entropy as our loss function, which is formalized as follows:
\begin{equation}
\begin{split}
J(\theta)=-\frac{1}{n_d}[\sum_{i=1}^{n_d}\sum_{j=1}^{n_e}1\{y^{(i)}=e_j\}\log p(e_i|d)].
\end{split}
\end{equation}
$n_d$ and $n_e$ are the number of dialogue and emoji. $1\{y^{(i)}=e_j\}$ equals $1$ only if the reply sentence of the $i$-th dialogue have the $j$-th emoji, and otherwise equals $0$.

\subsection{Optimization and Implementation Details}

The overall emoji classification models can be considered as a parameter set $\mathbf{\theta} = (\mathbf{X}, \mathbf{W}, \mathbf{U}, \mathbf{b})$, in which $\mathbf{X}$ represents the embeddings of all words. $\mathbf{W}$ and $\mathbf{U}$ stand for the projection matrices, while $\mathbf{b}$ stands for the bias vector.

Both LSTM and H-LSTM models are optimized with AdaDelta \cite{Zeiler2012adadelta}, with chain rule applied to update all parameters. All parameters are initialized randomly. To avoid overfitting, we implement a dropout layer before the softmax layer. We also adopt the early stop strategy which will terminate the training process when the error rate on validation set doesn't decrease in a few iterations. We implement all models with Theano \cite{bergstra2010theano,bastien2012theano}. For the consideration of efficiency, we utilize GPU to accelerate the training process.

\section{Experiments}

\subsection{Dataset}

In this paper, we utilize the Weibo2015 dialogues in Chinese as the original dataset. We first select $10$ emojis with relative high frequencies as our labels in classification. Afterwards, we extract all dialogues whose reply sentences contain only one of those emojis. Since the dialogue topics in Weibo2015 usually change frequently, we constraint the max length of a dialogue to be $4$ sentences to alleviate the non-essential information in long-term dependencies. We also attempt to balance the instances of different emojis, making each emoji to have nearly the same amount of dialogues.

In preprocessing, we implement some data cleaning procedures on these extracted raw dialogues. We wipe out all Weibo user names, quotes and transmission information, and also remove all emojis for fair predictions. To balance both effectiveness and efficiency, we construct the word dictionary according to the word frequency, and those words whose frequencies are less than $30$ are considered to be out-of-vocabulary words (OOVs). We further discard the dialogues that contain the sentences whose lengths are more than $50$ words or OOVs percentages are more than $25\%$. All dialogues after data preprocessing are randomly split into train, validation and test set. The statistics of dataset are listed in Table \ref{tab. 1}.

\begin{table}[!h]
\center
\small
\caption{\label{tab. 1} Statistics of the dataset}
\begin{tabular}{ccccc}
 \toprule
  Dataset & \#Emoji & \#Train & \#Valid & \#Test\\
 \midrule
  Weibo2015 & 10 & 1,164,694 & 64,732 & 64,271 \\
 \bottomrule
\end{tabular}
\end{table}

\subsection{Experiment Settings}

Our three models are trained via AdaDelta, with the decay constant $\rho=0.95$. The dropout ratio $\gamma$ is set to be $0.5$. We select the dimension of word embeddings $n_x$ and the dimension of hidden embeddings $n_h$ among $\{64, 128, 256, 384, 512\}$, and the mini-batch size $B$ among $\{16, 32, 64, 128\}$. The optimal configurations of our models are: $n_x=n_h=384$, $B=128$. As what has been stated above, the max length of dialogues is $4$ while the max length of sentences is $50$.

For baselines, we implement a bag-of-words model (BOW) which takes logistic regression as the classifier and TFIDF as features. For fair comparisons, the experimental setting of the baseline is the same as those of our models stated above. In the following sections, S-BOW and S-LSTM represent their corresponding models with single-turn dialogue inputs, F-BOW and F-LSTM represent those with multi-turn dialogue inputs, while H-LSTM represents the hierarchical LSTM model also with multi-turn dialogue inputs.

\begin{figure*}[!htbp]
\centering
\centering
\subfigure[P@1]{
\label{Fig.sub.1}
\includegraphics[width=0.28\textwidth]{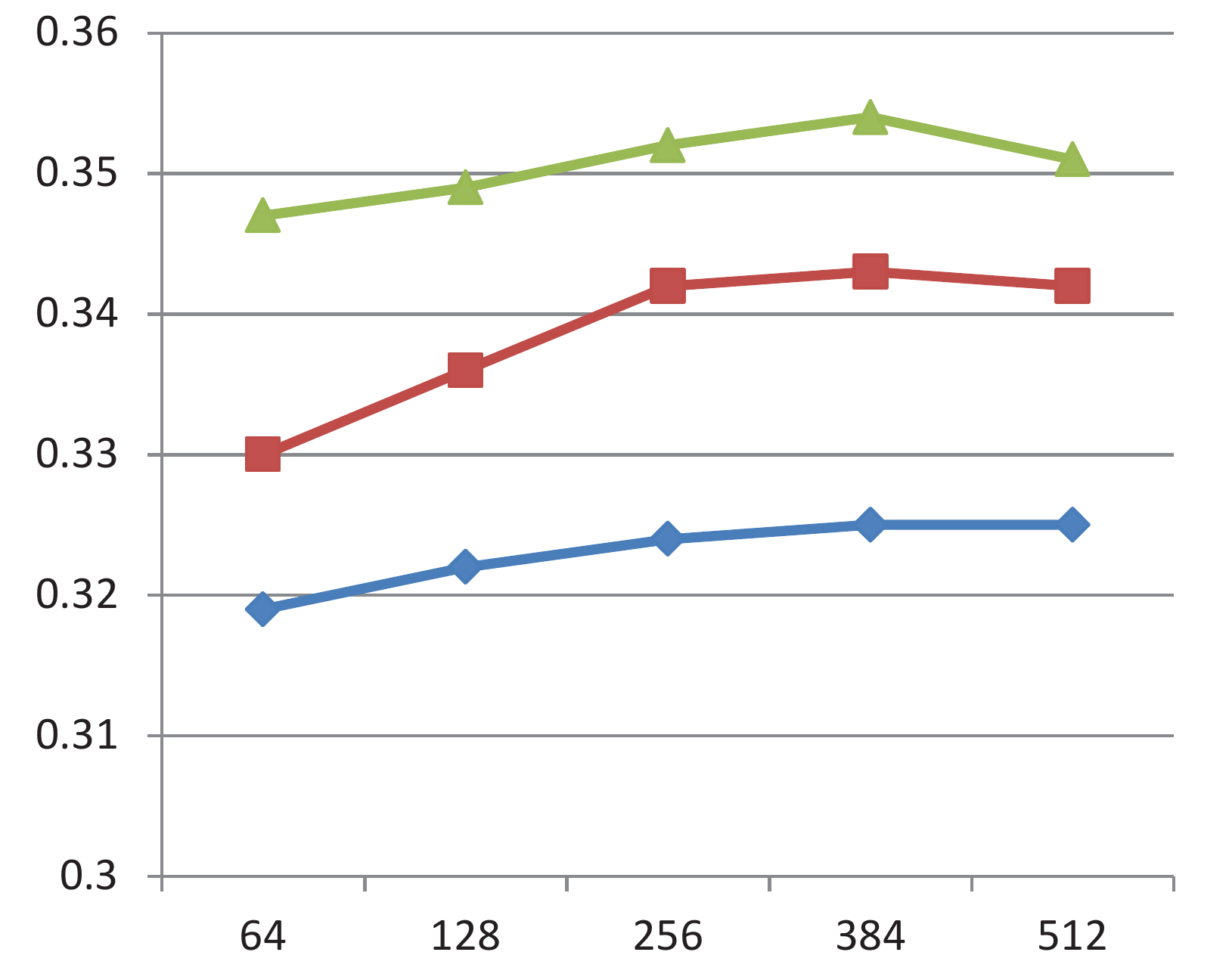}}
\subfigure[P@3]{
\label{Fig.sub.2}
\includegraphics[width=0.28\textwidth]{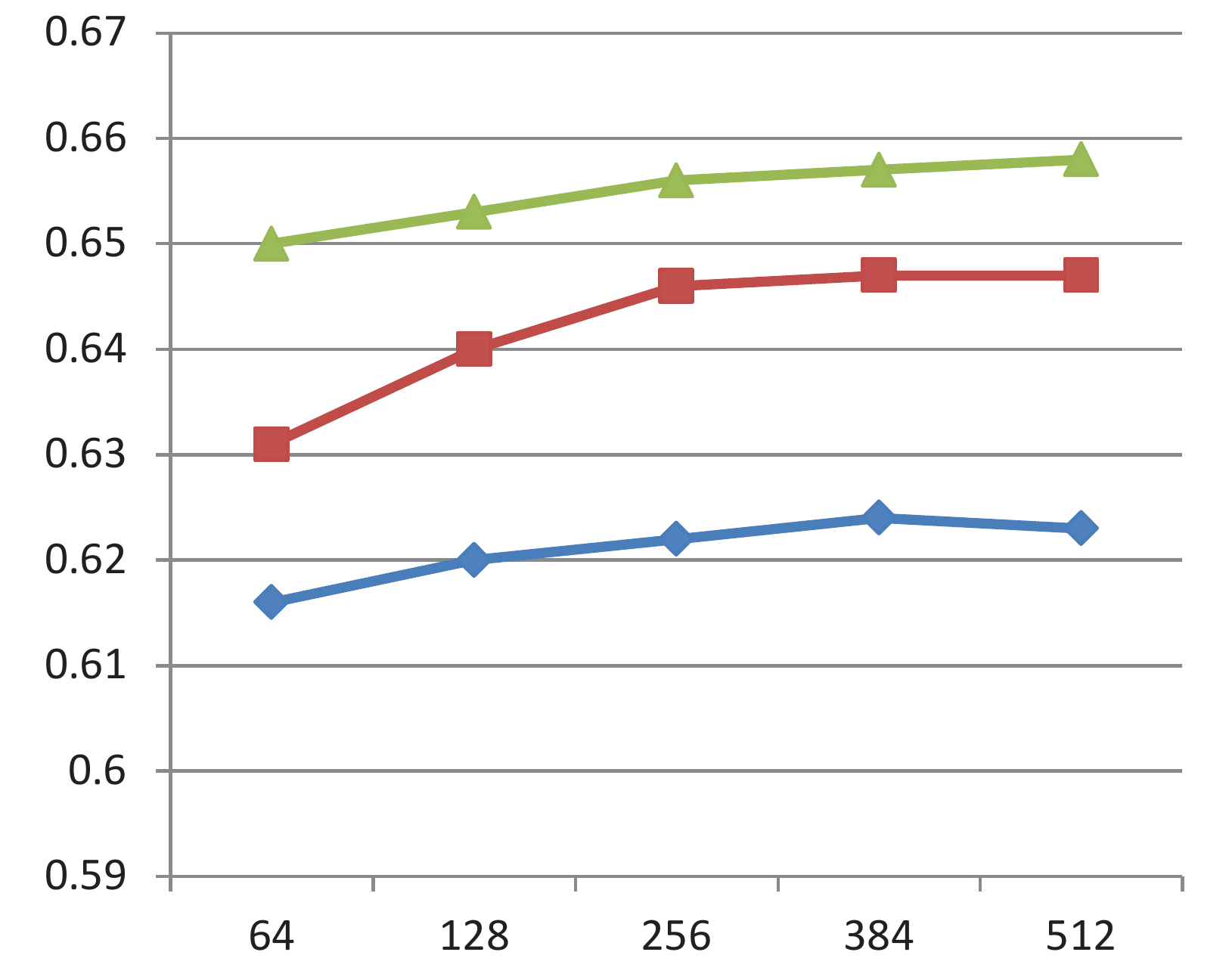}}
\subfigure[MRR]{
\label{Fig.sub.3}
\includegraphics[width=0.28\textwidth]{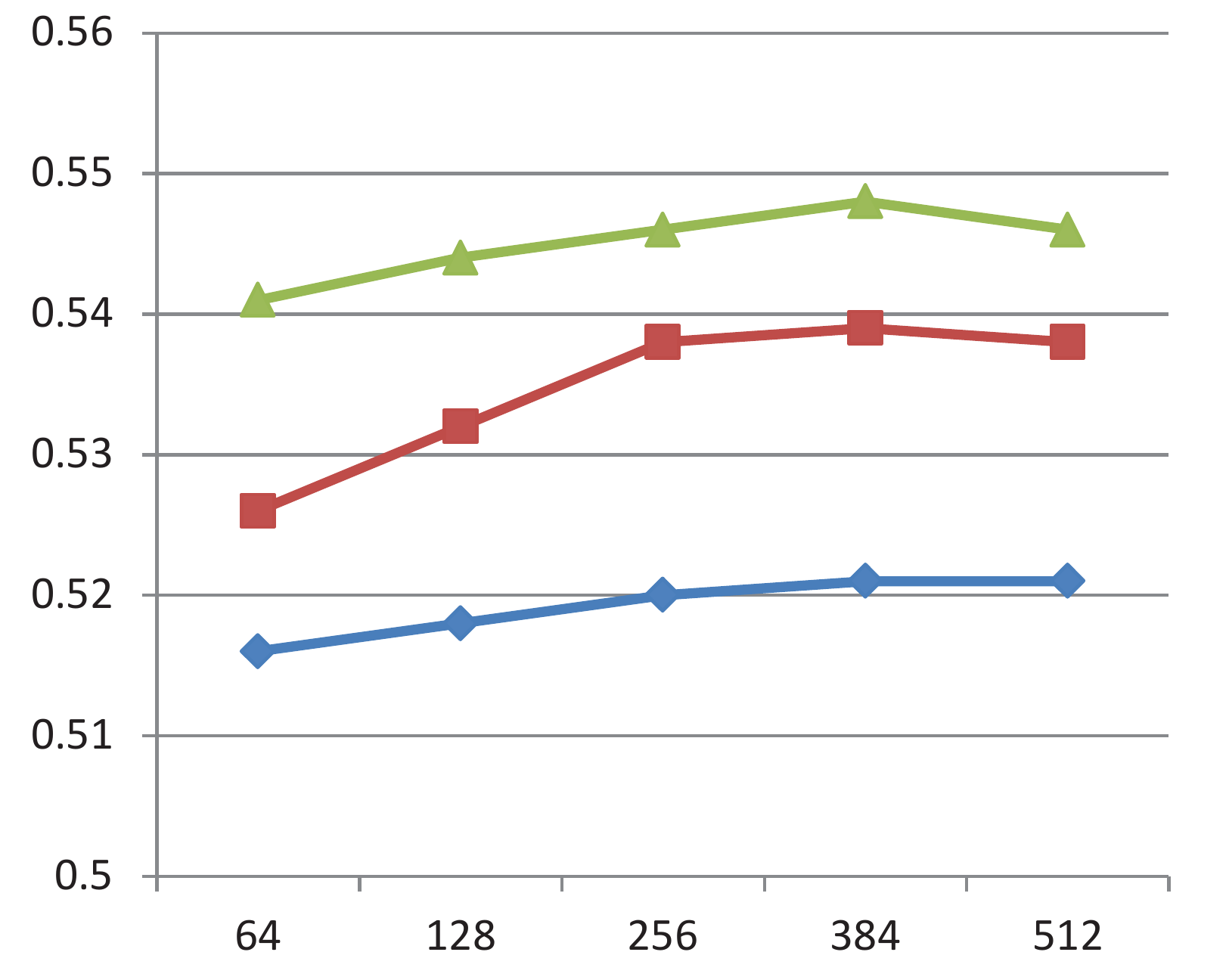}}
\includegraphics[width=0.07\textwidth]{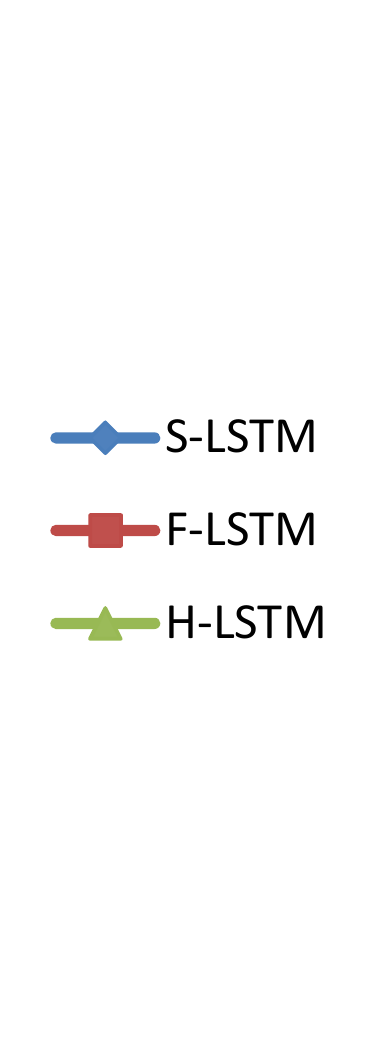}
\caption{Evaluation results with different dimensions.}
\label{fig. 4}
\end{figure*}

\subsection{Emoji Classification}

The task of emoji classification aims to predict the appropriate emojis according to the reply sentences or the whole dialogues. It could be utilized in real-world dialogue systems when you are going to give some emojis with your generated reply sentences to make the communication more lively.

\subsubsection{Evaluation Protocal}

We implement two metrics to evaluate our models: (1) the precision in top $k$ emoji candidates (P@k), and (2) the mean reciprocal rank (MRR). P@k is one of the most common evaluation metrics widely utilized in classification tasks, which directly indicates the classification accuracies in top $k$ candidates. Moreover, P@k reflects the capability of real-world dialogue system generating emojis with reply sentences. In experiments, we report P@1, P@3 and MRR for evaluation. MRR is an evaluation metric concentrating on not the prediction probabilities but the prediction ranks. The evaluation metric of MRR reflects the overall qualities on the task of emoji classification.

\subsubsection{Experimental Results}

The experimental results on emoji classification are demonstrated in Table \ref{tab. 2}. From the results we can observe that:

(1) All our models with multi-turn dialogue inputs significantly outperform all baselines on both evaluation metrics including P@k and MRR. It indicates that the contextual information located in conversations is of great significance for emoji classification, which has been successfully embedded into the dialogue representations via the LSTM-based encoders.

(2) The H-LSTM model achieves the best performance among all models, which confirms the improvements introduced by the hierarchical structure in conversations when constructing dialogue representations. The H-LSTM model first learn the meanings of all sentences, and then attempts to understand the whole dialogue via these learned sentences. On the contrary, the F-LSTM model merely regards each dialogue as a long flattened sentence, while it is still challenging for LSTM modeling too much long-term dependencies.

(3) The F-LSTM model performs better compared to the S-LSTM model, which confirms the significance of multi-turn information. However, considering the whole dialogues in BOW models will surprisingly harm the performances. It is because that the bag-of-words assumption can hardly model long-term memories well, and thus may be challenging in understanding current emotions.

(4) The performances on emoji classification still seem to be far from perfectness. It is because that the use of emojis in real-world dialogues are more casual and unformatted compared to other classification tasks such as social labelling or sentiment analysis. That is to say, different emojis could all fit well in the same dialogues, and the real-world performances of our model are much better than the digits shown in evaluation. In case study, we will further discuss this phenomenon.

\begin{table}[!h]
\center
\small
\caption{\label{tab. 2} Evaluation results on emoji classification}
\begin{tabular}{p{1.5cm}<{\centering}p{1.25cm}<{\centering}p{1.25cm}<{\centering}p{1.25cm}<{\centering}}
 \toprule
  Method & P@1 (\%) & P@3 (\%) & MRR (\%)\\
 \midrule
  S-BOW & 29.6 & 57.9 & 49.1\\
  F-BOW & 24.6 & 51.3 & 44.3\\
 \midrule
  S-LSTM & 32.5 & 62.4 & 52.1\\
  F-LSTM & 34.3 & 64.7 & 53.9\\
  H-LSTM & \textbf{35.4} & \textbf{65.7} & \textbf{54.8}\\
 \bottomrule
\end{tabular}
\end{table}

For further comparisons, we report the P@1 results on different emoji categories. From Table \ref{tab. 3} we could observe that: (1) The evaluation results on different emojis show different performances, which implies that there are indeed existing emojis that are more confusing and harder to be predicted than other emojis. (2) Emojis such as \emph{heart} and \emph{angry} are relatively easier to be predicted, since this kind of emojis is more straightforward and is usually used in constrained contexts. (3) On the contrary, predicting emojis such as \emph{tears of joy} and \emph{thinking} are more challenging, for these emojis are more ambiguous and complicated. For instance, the emoji \emph{tears of joy} is usually used to express the compounded feeling mixed with slight sad, helpless and embarrassed. Such compounded emotions could be smoothly replaced by other emojis like \emph{cry} or \emph{nervous} according to the contexts. (4) H-LSTM has advantages over S-LSTM almost on every emoji category, especially on those more complicated emojis. It confirms that the information in multi-turn dialogue indeed helps understanding the emotions of more complicated dialogues.

\begin{table}[!h]
\center
\small
\caption{\label{tab. 3} Evaluation results of P@1 on different emojis}
\begin{tabular}{p{0.8cm}<{\centering}p{2.0cm}<{\centering}p{1.25cm}<{\centering}p{1.25cm}<{\centering}}
 \toprule
  Emoji & Definition & S-LSTM & H-LSTM\\
 \midrule
  \includegraphics[width=0.017\textwidth]{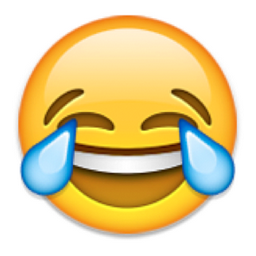}& \emph{tears of joy} & 16.5 & 21.6 \\
  \includegraphics[width=0.017\textwidth]{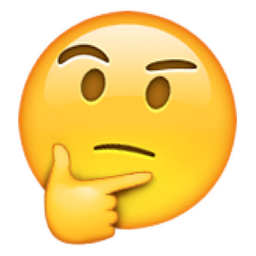}& \emph{thinking} & 21.6 & 22.7 \\
  \includegraphics[width=0.017\textwidth]{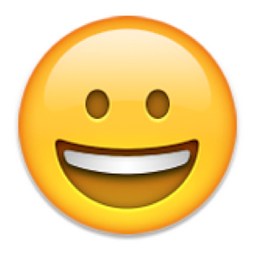}& \emph{laugh} & 17.5 & 24.1 \\
  \includegraphics[width=0.017\textwidth]{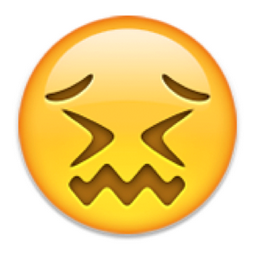}& \emph{nervous} & 23.2 & 27.1 \\
  \includegraphics[width=0.017\textwidth]{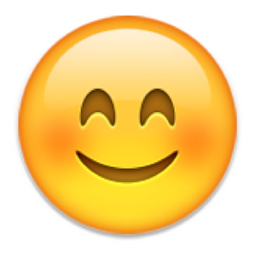}& \emph{shy} & 23.5 & 28.5 \\
  \includegraphics[width=0.017\textwidth]{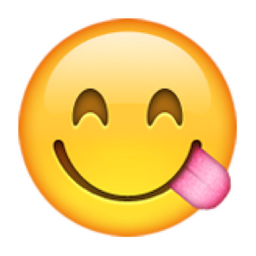}& \emph{delicious} & 33.1 & 32.7 \\
  \includegraphics[width=0.017\textwidth]{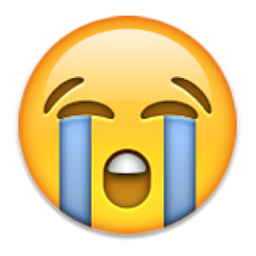}& \emph{cry} & 35.6 & 38.9 \\
  \includegraphics[width=0.017\textwidth]{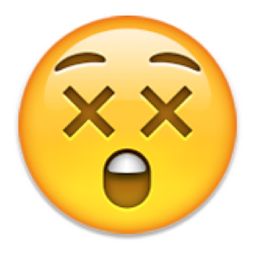} & \emph{astonished} & 46.6 & 47.4 \\
  \includegraphics[width=0.017\textwidth]{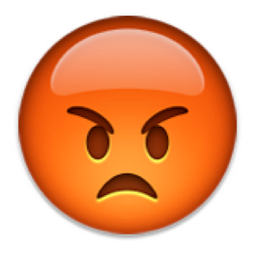}& \emph{angry} & 49.3 & 51.0 \\
  \includegraphics[width=0.017\textwidth]{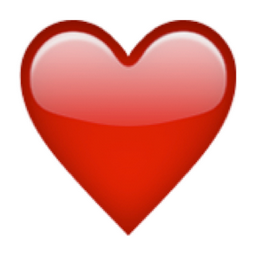}& \emph{heart} & 60.3 & 62.2 \\
 \bottomrule
\end{tabular}
\end{table}

\begin{CJK*}{UTF8}{gbsn}
\begin{table*}[!htbp]
\center
\small
\caption{\label{tab. 4} Examples of different models on emoji classification}
\begin{tabular}{p{1.0cm}<{\centering}p{10cm}p{1.2cm}<{\centering}p{1.2cm}<{\centering}p{1.2cm}<{\centering}}
 \toprule
  No. & Dialogue & S-LSTM & H-LSTM & Answer\\
 \midrule
  1 & A: 别哭了出去吃！\quad (Stop crying, and let's hang out for eating!)& & &\\
    & B: 去哪吃 \quad (To where?) & & &\\
    & A: 我之前收藏了一天关于宁波吃的的链接。随便找一家！\quad (I've collected lots of recommendations on eating in Ning Bo, we can choose from them!)& & &\\
    & B: 好呀好呀你啥时候有空 \quad (Great! When will you be free?)& \emph{shy} & \emph{delicious} & \emph{delicious}\\
  \midrule
  2 & A: 太过分了啊啊啊啊 \quad (It's so unacceptable!)& & &\\
    & B: 生气啊啊啊啊 \quad (I'm really angry!)& & &\\
    & A: 你生谁气 \quad (Who are you mad at?)& & &\\
    & B: 那个提香蕉的 \quad (The person who mentioned bananas!)& \emph{delicious} & \emph{nervous} & \emph{nervous}\\
  \midrule
  3 & A: 芭比娃娃一样?? \quad (Just like a barbie doll?)& & &\\
    & B: 太好看 \quad (It's so beautiful!)& & &\\
    & A: 哈哈哈谢谢 \quad (LOL, thank you!)& & &\\
    & B: 等等你的短发呢 \quad (Wait! Where is your short hair?)& \emph{thinking} & \emph{shy} & \emph{thinking}\\
  \midrule
  4 & A: 越画越好 \quad (Your paintings are getting better since you draw more.)& & &\\
    & B: 谢谢姐姐鼓励，画画真的让人开心 \quad (Thanks for your encourage, my sister. Drawing really makes me happy.)& & &\\
    & A: 是，这是个很好的爱好 \quad (I agree, that's a good hobby.)& & &\\
    & B: 跳舞也是 \quad (And so is dancing.)& \emph{laugh} & \emph{heart} & \emph{shy}\\
 \bottomrule
\end{tabular}
\end{table*}
\end{CJK*}

\subsection{Parameters Analysis}

The performances on emoji classification change with different parameter settings. In Fig. \ref{fig. 4}, we show a series of classification results with different word and hidden state dimensions to quantify the parameter sensitivities. The horizontal axis stands for the embedding dimensions, while the vertical axis represents the percentage of corresponding evaluation metrics. For better demonstrations, we assure that the dimensions of all hidden states and word embeddings are equal in each parameter setting.

From the results we can observe that: (1) All three models are sensitive with different word and hidden states dimensions. As the dimension increases, the performances of all models on both evaluation metrics will first get better and then remain stable or even get worse. The best performance appears when $n_x=n_h=384$ for all models. (2) The H-LSTM model significantly outperforms other LSTM models in all parameter settings, which confirms the robustness of H-LSTM in emoji classification.

\subsection{Case Study}

In this section, we give some representative examples on emoji classification for further discussions on the advantages as well as limitations in our models. The dialogues and their emojis for the reply sentences are demonstrated in Table \ref{tab. 4}.

In the first case, the reply sentence is a simple invitation, with the detailed information of the invitation located in the conversation above. H-LSTM predicts the correct emoji \emph{delicious} by considering the whole dialogue, while S-LSTM fails to fully understand the invitation due to the limited information in reply sentence. The same situation appears in the second case, in which S-LSTM misunderstands the topic of the whole dialogue strongly influenced by the \emph{banana} in reply sentence, and thus gives a \emph{delicious} prediction. However, the long-term dependencies won't always work since the conversation topic usually changes rapidly on real-world social platforms. In the third case, the central topic of the whole dialogue is about praises on a new makeup. Unfortunately, the emotion changes in the last sentence. S-LSTM concentrates on the short-term information in the reply sentence and get the right prediction, while H-LSTM hasn't forgotten the long-term memories and thus makes a mistake. Finally, we demonstrate a case where both S-LSTM and H-LSTM generate wrong emojis. This case demonstrates the difficulties in real-world emoji recommendation systems. Our models could learn that the emotion in this dialogue is rather positive, while it's extremely hard to go further and give detailed predictions. Emojis including \emph{laugh}, \emph{heart} and \emph{shy} all fit well in this dialogue, since those emojis could express correct emotions from different aspects and the answer is not unique. Making specifically correct predictions needs profound understandings in both linguistics and human behaviours, which is still challenging for us to achieve.

\section{Conclusion and Future Work}

In this paper, we focus on a novel task named emoji recommendation in multi-turn dialogue systems. We utilize hierarchical long short-term memory network to encode the contextual information located in conversations, and then give the appropriate predictions according to the learned dialogue representations. Experimental results and further discussions indicate that our method is capable of modeling contextual information for emoji classification.

We will explore the following research directions in future: (1) It is still challenging for our models to distinguish similar and interchangeable emojis according to minor differences in dialogues. More sophisticated models that is specifically designed for the scenario of dialogue system will help to address this issue. (2) The performances of emoji classification could be enhanced with better personalization and customization, which we will explore in future. (3) In this paper, we merely simplify the emoji recommendation task as a classification. We will explore more flexible emoji recommendation methods in dialogue systems with emoji positions and coherence into consideration, making communications more natural and lively.

\bibliographystyle{named}
\bibliography{reference}

\end{document}